\documentclass[letterpaper]{article} 
\usepackage{aaai2027}  
\usepackage[hyphens]{url}  
\usepackage{graphicx} 
\usepackage{natbib}  
\usepackage{caption} 
\usepackage{algorithm}
\usepackage{algorithmic}

\usepackage{newfloat}
\usepackage{listings}
\usepackage{microtype}
\usepackage{graphicx}
\usepackage{subcaption}
\usepackage{booktabs} 

\usepackage{url}
\usepackage{indentfirst}
\usepackage{algorithm}
\usepackage{algorithmic}
\usepackage{amsmath,amsfonts}
\usepackage{algorithmic}
\usepackage{array}
\usepackage{textcomp}
\usepackage{url}

\usepackage{epsfig}
\usepackage{graphicx}
\usepackage{amsmath}
\usepackage{multirow}
\usepackage{booktabs}
\usepackage{amsthm,amsmath}
\usepackage{amsmath}
\usepackage{cases}
\usepackage{bm}
\usepackage{color}
\usepackage{mathrsfs}
\usepackage[most]{tcolorbox}
\usepackage[table]{xcolor}
\usepackage{mdframed}
\usepackage{tcolorbox}
\usepackage{enumitem}
\usepackage{algorithmic}
\usepackage{amsmath}
\usepackage{graphicx} 
\DeclareCaptionStyle{ruled}{labelfont=normalfont,labelsep=colon,strut=off} 
\floatstyle{ruled}
\newfloat{listing}{tb}{lst}{}
\floatname{listing}{Listing}

\usepackage{booktabs}

\nocopyright

\title{S-GRPO: Unified Post-Training for Large Vision-Language Models}
\author{
    Yuming Yan\textsuperscript{\rm 1,\rm 2}\thanks{ORCID: 0009-0005-4596-1600},
    Kai Tang\textsuperscript{\rm 1},
    Sihong Chen\textsuperscript{\rm 1},
    Ke Xu\textsuperscript{\rm 1},
    Jialiang Yang\textsuperscript{\rm 1,\rm 3},
    Dan Hu\textsuperscript{\rm 1},
    Xiaoyu Wang\textsuperscript{\rm 1},
    Qun Yu\textsuperscript{\rm 1},
    Pengfei Hu\textsuperscript{\rm 1}
}
\affiliations{
    \textsuperscript{\rm 1}Tencent, Shenzhen, Guangdong, China\\
    \textsuperscript{\rm 2}Zhejiang University, Hangzhou, Zhejiang, China\\
    \textsuperscript{\rm 3}Tsinghua University, Beijing, China\\
    yominyan@tencent.com, xxx@tencent.com, xxx@tencent.com, xxx@tencent.com, xxx@tencent.com, xxx@tencent.com, xxx@tencent.com, xxx@tencent.com, xxx@tencent.com
}

\begin{document}

\maketitle

\begin{abstract}
Current post-training methodologies for adapting Large Vision-Language Models (LVLMs) generally fall into two distinct paradigms: Supervised Fine-Tuning (SFT) and Reinforcement Learning (RL). Despite their prevalence, both approaches suffer from significant inherent inefficiencies when applied in isolation. SFT forces the model's generation along a single, deterministic expert trajectory, often inducing catastrophic forgetting of general multimodal capabilities due to distributional shifts. Conversely, pure RL explores multiple generated trajectories but frequently encounters optimization collapse—a "cold-start" problem where an unaligned model fails to spontaneously sample any domain-valid trajectory in sparse-reward visual tasks. In this paper, we propose Supervised Group Relative Policy Optimization (S-GRPO), a unified post-training framework that seamlessly integrates the definitive guidance of imitation learning into the multi-trajectory exploration of preference optimization. Tailored for direct-generation visual tasks, S-GRPO introduces Conditional Ground-Truth Trajectory Injection (CGI). When a binary verifier detects a complete exploratory failure within a sampled group of trajectories, CGI dynamically injects the verified ground-truth trajectory into the candidate pool. By assigning a deterministic maximal reward to this injected anchor, S-GRPO enforces a strictly positive signal within the group-relative advantage estimation. This mechanism reformulates the supervised learning objective as a high-advantage component of the policy gradient, compelling the model to dynamically balance between exploiting the expert trajectory and exploring novel visual concepts. Theoretical analysis and empirical results demonstrate that S-GRPO gracefully bridges the gap between SFT and RL, drastically accelerates convergence, and achieves superior domain adaptation while flawlessly preserving the base model's general-purpose capabilities.
\end{abstract}


\section{Introduction}
Large Vision-Language Models (LVLMs) \citep{Qwen2.5-VL,Qwen2-VL, Qwen-VL, wang2025internvl3_5, zhu2025internvl3, chen2024expanding, wang2024mpo,gao2024mini, liu2024llavanext, liu2023improvedllava, qwen3.5,Qwen3-VL,chen2024internvl} unify visual perception with natural language understanding, and alignment research is increasingly focused on adapting these foundational models to highly specialized domains—such as medical image diagnostics, complex geometric problem-solving, and fine-grained industrial visual inspection—where the model must adhere to rigorous logical constraints and domain-specific rubrics beyond generic perception. Prevailing post-training techniques for such domain adaptation typically fall into two categories: Supervised Fine-Tuning (SFT-like methods) \citep{Ouyang2022TrainingLM,Hu2021LoRALA,Li2021PrefixTuningOC} and Reinforcement Learning (RL-like methods) \citep{Shao2024DeepSeekMathPT,Yu2025DAPOAO,Zheng2025GroupSP,Zhang2025OnPolicyRM,Xi2024TrainingLL,Deng2025BoostingTG,Wu2025GCPOWC,Zhang2025ScafGRPOSG,Xiao2025FastSlowTG}.
\begin{figure}
  \includegraphics[width=0.47\textwidth]{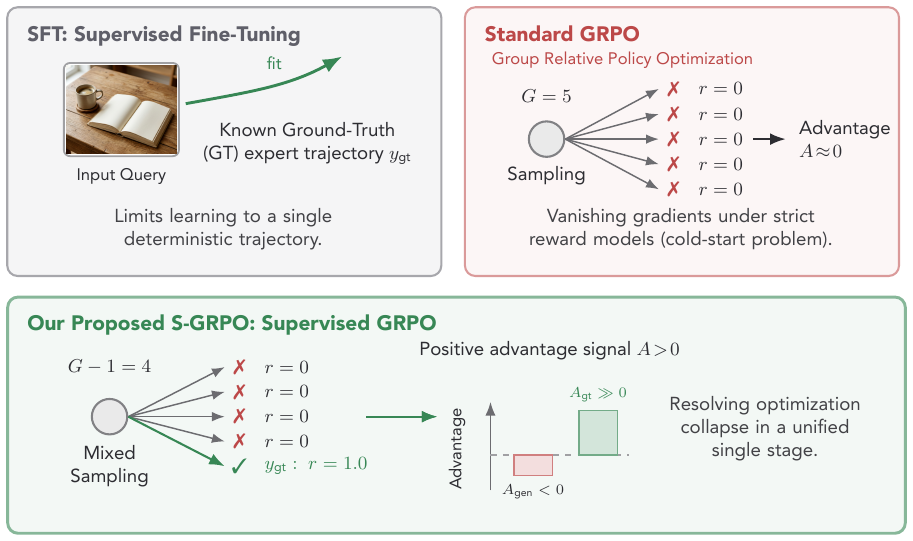}
  \caption{Conceptual comparison of post-training paradigms. \textbf{Top Left:} SFT limits learning to a single deterministic trajectory. \textbf{Top Right:} Standard GRPO explores multiple trajectories but suffers from vanishing gradients ($A \approx 0$) when all generated samples fail under strict reward models. \textbf{Bottom:} Our proposed S-GRPO conditionally injects the ground-truth trajectory into the candidate group. This guarantees a positive advantage signal ($A > 0$), effectively resolving the cold-start problem and preventing optimization collapse in a unified single stage.}
  \label{fig:motivation}
\end{figure}
Although both paradigms are widely adopted, as shown in Fig. \ref{fig:motivation}, each suffers from structural limitations resulting from how they manage the autoregressive generation process. SFT \citep{Ouyang2022TrainingLM} utilizes maximum likelihood estimation (MLE) to force the model along a \textit{single, deterministic ground-truth trajectory}. While highly efficient for injecting specific domain knowledge, this heavy reliance on strict teacher-forcing introduces a significant distributional shift, often triggering the "Alignment Tax" \citep{Lin2023MitigatingTA}—a form of catastrophic forgetting where the model's broad, general-purpose multimodal competencies are overwritten by the narrow target distribution.

In stark contrast, RL algorithms—such as Group Relative Policy Optimization \citep{Shao2024DeepSeekMathPT} (GRPO)—refine the policy by sampling \textit{multiple diverse trajectories} against a reward function. However, directly applying RL to an unaligned base model exposes a severe "cold-start" problem \citep{Wu2025GCPOWC}: in specialized visual domains evaluated by strict binary verifiers, the base model struggles to spontaneously generate accurate trajectories, and if all sampled trajectories in a group are incorrect, the reward variance collapses to zero, causing vanishing gradients and stalled learning. Practitioners thus face a dilemma: SFT causes catastrophic forgetting, while pure RL suffers optimization collapse from reward sparsity.

We argue that this dichotomy is artificial: mathematically, SFT can be viewed as an RL process where the ground-truth trajectory is exclusively assigned a maximal reward, suggesting the transition from supervised to reinforcement learning should be a continuous spectrum rather than disjointed stages. By treating the ground-truth trajectory as an active participant in the preference comparison—an "anchor" with guaranteed high reward—we leverage the contrastive nature of policy gradients to unify expert imitation with autonomous multi-trajectory exploration within a single optimization landscape.

Based on this insight, we propose Supervised Group Relative Policy Optimization (S-GRPO), a unified post-training framework tailored for LVLMs, introducing a dynamic mechanism: \textit{Conditional Ground-Truth Trajectory Injection (CGI)}. During online sampling, if a binary verifier detects that the policy has failed to generate any valid trajectory within an exploratory group, S-GRPO conditionally injects the verified ground-truth trajectory into the candidate pool, and the relative advantage is computed against this mixed group. The injected trajectory acts as a high-advantage anchor guiding the gradient toward the correct multimodal distribution; as the model improves, the CGI mechanism gracefully phases out, allowing training to transition from supervised bootstrapping to pure self-exploratory reinforcement learning.
Our main contributions are summarized as follows:
\begin{itemize}
\item \textbf{Unified Post-Training Framework}: We formalize S-GRPO, an algorithm that elegantly integrates the objectives of SFT and preference optimization, eliminating the need to choose between isolated paradigms and mitigating catastrophic forgetting in LVLMs.
\item \textbf{Mechanism Design}: We propose the Conditional Ground-Truth Trajectory Injection (CGI) strategy, which dynamically solves the exploration cold-start problem in binary-verifier settings by guaranteeing positive learning signals within multi-trajectory updates.
\item \textbf{Empirical Performance}: We demonstrate that S-GRPO significantly accelerates convergence in sparse-reward visual domains and achieves superior domain-specific performance while preserving broad general capabilities compared to other methods.
\end{itemize}

\section{Related Works}
\subsection{Large Vision Language Models (LVLMs)}
Large Vision-Language Models (LVLMs) \citep{Qwen2.5-VL,Qwen2-VL,Qwen-VL,wang2025internvl3_5,zhu2025internvl3,chen2024expanding,wang2024mpo,gao2024mini,liu2024llavanext,liu2023improvedllava,qwen3.5,Qwen3-VL,chen2024internvl,kimiteam2026kimik25visualagentic} have driven remarkable advancements in cross-modal intelligence. This evolution unfolded across three key phases: early foundational architectures (e.g., CLIP \citep{Radford2021LearningTV}, ALBEF \citep{Li2021AlignBF}) bridged the modality gap for zero-shot transfer; subsequent models like LLaVA \citep{Liu2023VisualIT} elevated general capabilities via visual instruction tuning; and current state-of-the-art models—such as the InternVL series \citep{wang2025internvl3_5,zhu2025internvl3,wang2024mpo}, Qwen-VL family \citep{Qwen2.5-VL, Qwen2-VL, Qwen-VL,qwen3.5,Qwen3-VL}, and Kimi K2.5 \citep{kimiteam2026kimik25visualagentic}, push the boundaries through massive scaling, high-resolution encoders, and native long-context processing. Despite these capabilities, LVLMs still struggle with domain-specific adaptation (e.g., medical diagnostics), making post-training a critical direction for adapting these massive models efficiently without the prohibitive cost of retraining from scratch.

\subsection{Post-training for LVLMs}
Post-training techniques, primarily Supervised Fine-Tuning (SFT) and Reinforcement Learning (RL), have been central to adapting LVLMs \citep{kumar2025llmposttrainingdeepdive, chu2025sft, lai2025med, li2025drive}. SFT enables direct task-specific learning but often leads to catastrophic forgetting of pre-trained knowledge \citep{duan2024cityllava, dong2025scalable,chen2024efficiency}; parameter-efficient approaches like LoRA \citep{Hu2021LoRALA} and Prefix Tuning \citep{Li2021PrefixTuningOC} alleviate computational burdens but remain prone to overfitting.

RL-based methods instead optimize sequential decision-making. Unlike PPO \citep{Schulman2017ProximalPO}, recent alternatives such as DPO \citep{Rafailov2023DirectPO} and GRPO \citep{Shao2024DeepSeekMathPT} eliminate the need for a separate critic. Advanced frameworks (e.g., DAPO \citep{Yu2025DAPOAO}, GSPO \citep{Zheng2025GroupSP}) preserve general capabilities via diverse trajectory exploration, but implicitly assume non-trivial pre-existing domain knowledge; when domain expertise is limited, they suffer from reward sparsity and optimization collapse. The common ``SFT-then-RL'' pipeline \citep{Shao2024DeepSeekMathPT} attempts to alleviate this, but CHORD \citep{Zhang2025OnPolicyRM} shows that SFT disrupts pretrained structures, and subsequent RL often fails to recover domain adaptation.

To mitigate optimization collapse, curriculum-based approaches (e.g., R3 \citep{Xi2024TrainingLL}, Curr-ReFT \citep{Deng2025BoostingTG}) smooth the learning curve, GCPO \citep{Wu2025GCPOWC} and Scaf-GRPO \citep{Zhang2025ScafGRPOSG} address the ``learning cliff'' via golden-answer or hierarchical-hint injection, and FAST-GRPO \citep{Xiao2025FastSlowTG} mitigates ``overthinking'' via difficulty-aware length penalties—though these often rely on decoupled multi-stage pipelines, external teacher models, or Chain-of-Thought-specific designs.

Closer to our setting, RCPA \citep{Yan2026ReinforcedCP} addresses optimization collapse in non-deep-thinking vision tasks via Curriculum Progress Perception (CPP) intrinsic prefix-injection, bootstrapping domain competence directly within the on-policy RL loop; however, RCPA still necessitates a carefully designed curriculum schedule and explicit manipulation of the generation prefix, adding algorithmic complexity and restricting the model's free exploration.

Building upon the insights of RCPA, S-GRPO offers a more elegant, genuinely unified single-stage solution: instead of relying on predefined curriculum progression or prefix-injection, it utilizes Conditional Ground-Truth Trajectory Injection (CGI), dynamically triggered by group-level exploratory failure. By anchoring the policy gradient with verified expert trajectories only when strictly necessary, S-GRPO natively resolves extreme reward sparsity, bridging the SFT and RL paradigms without curriculum-design overhead while preserving the structural integrity and general capabilities of the underlying base LVLM.
\section{Preliminary}

\subsection{Problem Formulation}
Consider a pre-trained LVLM $\pi_{\theta_{\text{pre}}}$ with strong general-purpose multimodal capabilities, and a target domain-specific dataset $\mathcal{D}_{\text{target}} = \{(x_i, y_{gt,i})\}_{i=1}^N$, where $x_i = (v_i, q_i)$ is a multimodal query (image $v_i$ with task prompt $q_i$) and $y_{gt,i}$ is the ground-truth response. The objective is to adapt $\pi_{\theta_{\text{pre}}}$ into $\pi_{\theta}$ that performs well on $\mathcal{D}_{\text{target}}$ while maximally preserving general capabilities—the fundamental challenge of integrating novel domain knowledge without incurring catastrophic forgetting of prior knowledge.

\subsection{Standard Paradigms and Their Limitations}
\textbf{Supervised Fine-Tuning (SFT).} SFT adapts model parameters $\theta$ by maximizing the likelihood of expert demonstrations in the target dataset. Its objective is:
\begin{equation}
\mathcal{J}_{\text{SFT}}(\theta) = \mathbb{E}_{(x, y_{gt}) \sim \mathcal{D}_{\text{target}}} \left[ \sum_{t=1}^{|y_{gt}|} \log \pi_{\theta}(y_{gt,t} \mid x, y_{gt,<t}) \right].
\label{eq:sft_objective}
\end{equation}
While SFT effectively injects domain knowledge, it forces generation along a single deterministic trajectory; when target data differs substantially from pretraining data, this strict teacher-forcing causes catastrophic forgetting of the model's multimodal generalizability.

\textbf{Group Relative Policy Optimization (GRPO).} To mitigate SFT-induced forgetting, preference-based RL methods like GRPO align models via multi-trajectory exploration: given a query $x$, GRPO samples a group of $G$ trajectories $O_{gen}=\{o_1, \dots, o_G\}$ from the old policy $\pi_{\theta_{\text{old}}}$, a verifier assigns a scalar reward $r_i$ to each $o_i$, and the advantage estimate $\hat{A}_i$ is computed relative to the group peers:
\begin{equation}
\hat{A}_i = \frac{r_i - \text{mean}(\{r_1, \dots, r_G\})}{\text{std}(\{r_1, \dots, r_G\}) + \epsilon},
\label{eqn:grpo_adv}
\end{equation}
where $\epsilon$ is a small constant for numerical stability. The GRPO objective maximizes the expected clipped advantage while regularizing against a reference policy $\pi_{\theta_{\text{ref}}}$:
\begin{equation}
\begin{aligned}
    \mathcal{J}_{\text{GRPO}}(\theta) &= \mathbb{E}_{x, O_{gen} \sim \pi_{\theta_{\text{old}}}} \Bigg[ \frac{1}{G} \sum_{i=1}^G \min \Big( \rho_i(\theta) \hat{A}_i, \\
    &\quad \text{clip}\left(\rho_i(\theta), 1-\varepsilon, 1+\varepsilon\right) \hat{A}_i \Big) \Bigg] \\
    &\quad - \beta \mathbb{D}_{KL}(\pi_\theta \| \pi_{\theta_{\text{ref}}}),
\end{aligned}
\label{eqn:grpo_obj}
\end{equation}
where $\rho_i(\theta) = \frac{\pi_\theta(o_i|x)}{\pi_{\theta_{\text{old}}}(o_i|x)}$ is the importance sampling ratio, $\pi_{\theta_{\text{old}}}$ is the behavior policy used for sampling, and $\pi_{\theta_{\text{ref}}}$ is typically a frozen initial policy (e.g., the SFT model) that prevents excessive policy drift via the KL penalty.

\subsection{The Cold-Start Challenge and Optimization Collapse}
\label{ssec:cold_start}
While GRPO explores multiple trajectories effectively, it relies on the unaligned base policy spontaneously generating at least some correct trajectories within a group to establish a meaningful contrastive gradient.

However, in specialized visual domains evaluated by strict verifiers, the base model suffers from a severe \textit{cold-start problem}: the probability of spontaneously generating a valid response is infinitesimally small, so all $G$ sampled trajectories are likely to fail, resulting in uniformly zero rewards (e.g., $r_i = 0$ for all $i$). When this occurs, the group statistics collapse:
$$ \text{mean}(\{r_i\}) \approx 0, \quad \text{std}(\{r_i\}) \approx 0 $$
According to Eq. \ref{eqn:grpo_adv}, this leads to uninformative, near-zero advantage estimates ($\hat{A}_i \approx 0$). The contrastive signals vanish, causing an \textit{optimization collapse} where learning completely stalls. This fundamental limitation motivates our proposed unified approach, which actively injects supervisory signals into the RL exploration loop to dynamically break the optimization deadlock.
\subsection{Unified View of Post-training}
SFT and GRPO exist on a continuous spectrum. Setting $G=1$, $\hat{A}_i=r_i$, $o_i=y_{gt}$, and $\beta=0$ in the GRPO objective yields $L_{\mathrm{GRPO}} = -\mathbb{E}\left[ \frac{\pi_\theta(y_{gt}|x)}{\pi_{\theta_{\mathrm{old}}}(y_{gt}|x)} r \right]$. Applying a first-order Taylor expansion around $\theta_{\mathrm{old}}$ ($\theta \approx \theta_{\mathrm{old}}$):
$$\frac{\pi_\theta(y_{gt}|x)}{\pi_{\theta_{\mathrm{old}}}(y_{gt}|x)} \approx 1 + \log \pi_\theta(y_{gt}|x) - \log \pi_{\theta_{\mathrm{old}}}(y_{gt}|x)$$
Substituting this back, and dropping constants w.r.t. $\theta$, the objective reduces to $L_{\mathrm{GRPO}} = -\mathbb{E}\left[ r \log \pi_\theta(y_{gt}|x) \right]$. For $r=1$, this exactly reduces to the SFT loss: $L_{\mathrm{SFT}} = -\mathbb{E}\left[ \log \pi_\theta(y_{gt}|x) \right]$. Thus, SFT can be interpreted as a special case of policy-gradient optimization under constrained (single-trajectory) exploration, motivating a unified treatment of both paradigms within a single objective.
\section{Methodology}

\begin{figure*}[t]
  \centering
  \includegraphics[width=0.98\textwidth]{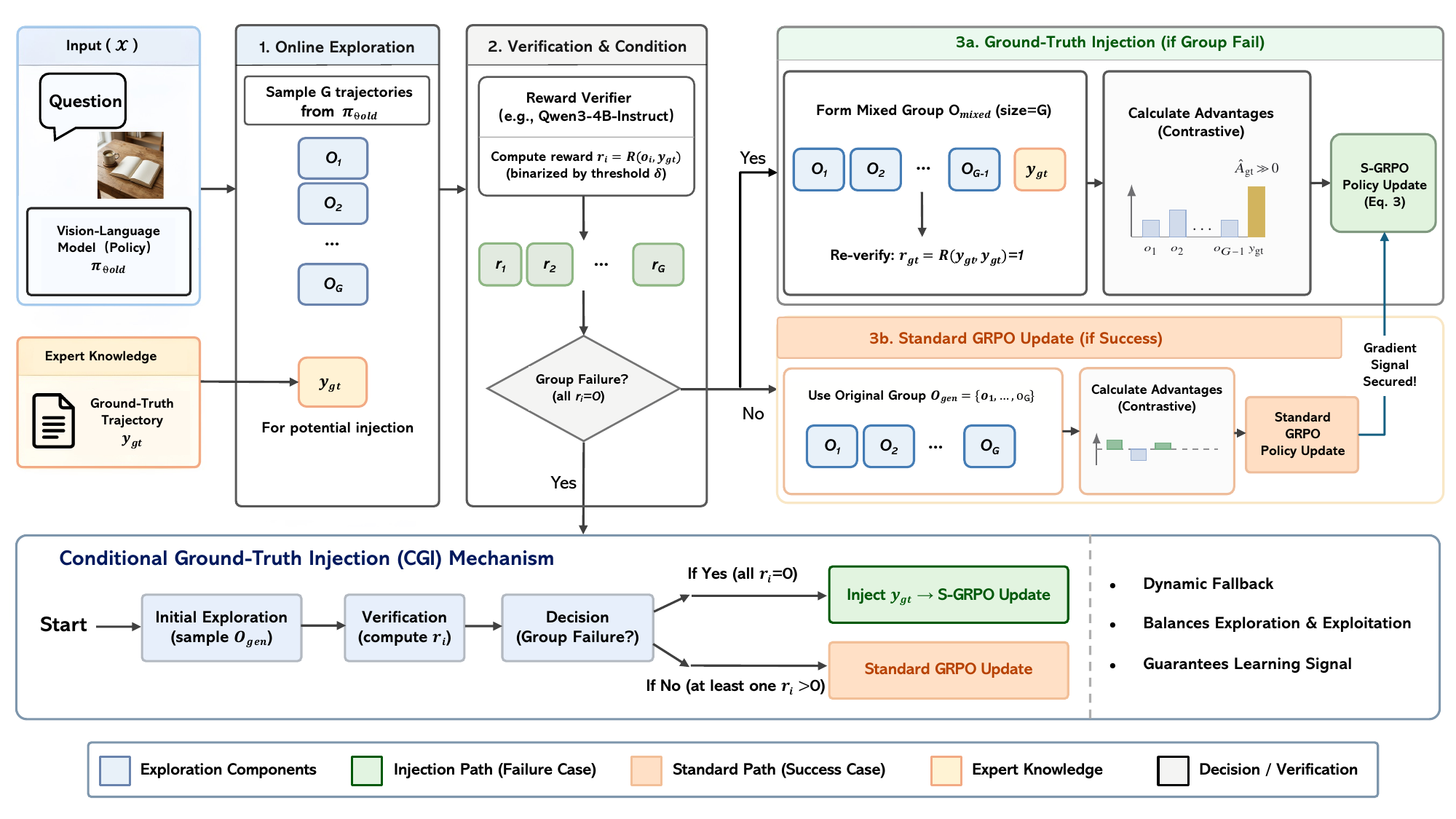}
\caption{Overview of the S-GRPO framework. The core innovation is the Conditional Ground-Truth Injection (CGI) mechanism: the current policy generates a group of candidate trajectories ($O_{gen}$) evaluated by a strict reward verifier; if it detects a complete exploratory failure (all $r_i=0$), CGI is triggered as a dynamic fallback (top path), injecting the expert trajectory ($y_{gt}$) to form a mixed group with a strongly positive advantage signal ($\hat{A}_{gt} \gg 0$) that prevents optimization collapse. If at least one valid trajectory is generated (bottom path), the framework defaults to a standard GRPO update, seamlessly balancing supervised imitation and autonomous exploration within a single stage.}
  \label{fig:method}
\end{figure*}

Building upon the limitations identified in the preliminary analysis, we introduce Supervised Group Relative Policy Optimization (S-GRPO). As illustrated in Fig. \ref{fig:method}, S-GRPO elegantly unifies the strengths of both SFT and GRPO into a single on-policy stage, resolving the cold-start challenge without decoupled training phases.

\subsection{Backbone and Reward Formulation}
Following \citep{Yan2026ReinforcedCP}, we focus on non-deep-thinking Visual Question Answering (VQA), and thus use GRPON (GRPO for non-deep-thinking models \citep{Yan2026ReinforcedCP}). A rule-based reward function or external evaluator $R(o, y_{gt})$ assigns scores to generated trajectories; specifically, we utilize Qwen3-4B-Instruct \citep{qwen3technicalreport} as our reward verifier for semantic consistency evaluation. The continuous similarity score returned by the verifier is then binarized using a threshold $\delta$ to provide clear guidance during advantage estimation:
\begin{equation}
\label{eq:reward}
    R(o, y; \delta) = 
    \begin{cases} 
    +1 & \text{if}~S(o, y) \ge \delta, \\
    0 & \text{otherwise}. 
    \end{cases} 
\end{equation}

\subsection{Supervised GRPO via Trajectory Injection}
To resolve the optimization collapse caused by reward sparsity, S-GRPO actively shapes the advantage estimation landscape by dynamically modifying the composition of the multi-trajectory sampling group.
\subsubsection{Mechanism: Ground-Truth Trajectory Injection}

Instead of relying solely on self-generated candidate trajectories, S-GRPO introduces an injection mechanism: for a given training query $x$ with known ground-truth expert trajectory $y_{gt}$, we sample $G-1$ exploratory trajectories from the current policy, $\{o_1, \dots, o_{G-1}\} \sim \pi_{\theta_{old}}(\cdot|x)$, and inject $y_{gt}$ to form a mixed candidate group of fixed size $G$: $O_{mixed} = \{o_1, \dots, o_{G-1}\} \cup \{y_{gt}\}$. Crucially, the injected trajectory $y_{gt}$ is assigned a guaranteed maximal reward $R_{max}=1.0$, acting as an intrinsic reference anchor, while the $G-1$ generated trajectories are evaluated by the verifier.

\subsubsection{The Unified S-GRPO Formulation}
Consider the mixed group $O_{mixed}$ in a severe cold-start scenario where all $G-1$ exploratory trajectories fail ($r=0$) and the injected expert trajectory succeeds ($r=R_{max}=1.0$). The group statistics intrinsically shift:
$$ \text{mean} = \frac{1}{G}, \quad \text{std} = \sqrt{\frac{G-1}{G^2}} \approx \frac{1}{\sqrt{G}} $$
The injection of a single high-reward anchor mathematically guarantees a non-zero standard deviation ($\text{std} > 0$), forcefully segregating the relative advantages: for generated (incorrect) trajectories, $\hat{A}_{gen} = \frac{0-(1/G)}{\text{std}} < 0$ (penalizing self-generated errors), while for the injected expert trajectory, $\hat{A}_{gt} = \frac{1-(1/G)}{\text{std}} \gg 0$ (strongly positive, imitating the reference). This stark contrast prevents gradient vanishing. The unified S-GRPO objective function is formally optimized over this mixed group:
\begin{equation}
\begin{aligned}
    \mathcal{J}_{\text{S-GRPO}}(\theta) &= \mathbb{E}_{x, O_{mixed} \sim \pi_{\theta_{\text{old}}}} \Bigg[ \frac{1}{G} \sum_{i=1}^G \min \Big( \rho_i(\theta) \hat{A}_i, \\
    &\quad \text{clip}\left(\rho_i(\theta), 1-\varepsilon, 1+\varepsilon\right) \hat{A}_i \Big) \Bigg] \\
    &\quad - \beta \mathbb{D}_{KL}(\pi_\theta \| \pi_{\theta_{\text{ref}}}),
\end{aligned}
\label{eqn:s_grpo_obj}
\end{equation}
where all notations and hyperparameters retain their definitions from Eq. \ref{eqn:grpo_obj}. Maximizing this objective naturally increases the log-probability of the expert trajectory (effectively executing an on-policy behavioral cloning step) while actively decreasing the probabilities of self-generated erroneous or low-quality trajectories.

\subsection{Conditional Ground-Truth Trajectory Injection (CGI)}
While baseline injection seamlessly mitigates the cold-start problem, unconditionally injecting the ground-truth trajectory throughout the entire training process may inadvertently reintroduce the strict teacher-forcing bias prevalent in SFT. To balance the exploration-exploitation trade-off, we propose a refined variant: \textit{Conditional Ground-Truth Trajectory Injection (CGI)}.

CGI employs the ground-truth trajectory $y_{gt}$ strictly as a dynamic fallback mechanism, triggered exclusively when the policy experiences a complete exploratory failure. For a query $x$, the process is as follows:

\begin{enumerate}
    \item \textbf{Initial Exploration:} We sample a full group of $G$ candidate trajectories strictly from the current policy: $O_{gen} = \{o_1, \dots, o_G\} \sim \pi_{\theta_{old}}(\cdot|x)$.
    \item \textbf{Verification and Failure Detection:} The verifier evaluates all $G$ trajectories to assign rewards $\{r_1, \dots, r_G\}$. We define a \textit{group failure} condition as the scenario where all generated trajectories fail to yield a positive reward (e.g., $\max(\{r_1, \dots, r_G\}) = 0$).
    \item \textbf{Conditional Fallback:} If the condition is met (complete failure), the model is trapped in a cold-start state: we discard one generated trajectory (the one with lowest generation probability) and replace it with $y_{gt}$, forming $O_{mixed}$, and the policy update proceeds using the unified S-GRPO formulation (Eq. \ref{eqn:s_grpo_obj}). Otherwise, if at least one generated trajectory receives a positive reward, the model has sufficient capability to learn from its own exploratory variance, so no injection occurs and the policy is updated using the standard GRPO objective (Eq. \ref{eqn:grpo_obj}) over $O_{gen}$.
\end{enumerate}
Unlike GCPO's unconditional injection (SFT-like teacher-forcing bias) and RCPA's curriculum-dependent prefix injection (manual design overhead), S-GRPO's CGI dynamically triggers ground-truth injection only when necessary, achieving a seamless transition from supervised bootstrapping to pure reinforcement learning in a single stage.
\section{Experiments}
In this section, we empirically validate the effectiveness of S-GRPO on Large Vision-Language Models. Our experiments are designed to address three key questions:
\textbf{(1)} Can S-GRPO successfully train a visual generation policy from a "cold-start" state where a strict binary verifier causes standard RL methods to collapse?
\textbf{(2)} How does the unified single-stage S-GRPO compare against the prevalent two-stage pipeline (SFT followed by GRPO) in terms of domain-specific performance and general capability retention?
\textbf{(3)} What is the impact of the Conditional Ground-Truth Injection (CGI) mechanism on learning dynamics?
\subsection{Experimental Setup}
\label{sec: setup}
\textbf{Benchmark Datasets.}
To evaluate S-GRPO's adaptability and performance, we conduct experiments on three benchmark datasets spanning diverse domains: image captioning, geometric problem-solving, and medical X-ray diagnostics.
\begin{itemize}[leftmargin=1em]
 \item COCO Caption~\citep{chen2015microsoft}: A widely used dataset for image captioning, containing 123,287 images. We use the original training and test splits of the dataset.
\item Geo170K~\citep{gao2025gllava}: A dataset for geometric problem-solving. It is divided into Phase 1 (non-deep-thinking, direct-answer tasks) and Phase 2 (deep-thinking, multi-step tasks). We use Phase 1, which contains 60,252 samples, with 57,252 for training and 3,000 for testing.
\item OpenI~\citep{demner2012design}: A chest X-ray diagnostic dataset with 6,423 images and corresponding radiological reports. The task is to generate concise and clinically accurate diagnostic descriptions directly from X-ray images. The training set consists of 5,423 images, while the test set includes 1,000 images.
\end{itemize}
These datasets enable us to evaluate S-GRPO's performance across both general domain (e.g., COCO Caption) and specific domain (e.g., Geo170K, OpenI) tasks, providing a comprehensive assessment of domain-adaptive capabilities.

\begin{table*}[tp]
\centering
{\renewcommand{\arraystretch}{0.85}
\resizebox{0.95\textwidth}{!}{
\begin{tabular}{clcccccccc}
\toprule
\multirow{2}{*}{Datasets} & \multicolumn{1}{c}{\multirow{2}{*}{Methods}} & \multicolumn{4}{c}{Domain-Specific Ability} & \multicolumn{4}{c}{General-Purpose Ability}\\ \cmidrule(r){3-6}\cmidrule(r){7-10} 
& \multicolumn{1}{c}{} & BLEU-1& ROUGE-L & CIDEr & SPICE & MMMU & MME & IFEval-P & IFEval-I \\ \midrule
\multirow{7}{*}{\shortstack{COCO\\Caption}} 
& BASE & 0.4457& 0.3672& 0.2259 & 0.1783& 0.5122 & \underline{2333.36} & 0.6211& 0.7038\\ \cmidrule{2-10} 
& PEFT & 0.3722& 0.3081& 0.1231 & 0.0862 & \textbf{0.6067} & \textbf{2448.67} & 0.5416& 0.6535\\
& FFT& \textbf{0.7581}& \textbf{0.5474}& \textbf{1.0172} & 0.2385& 0.4244 & 735.10 & 0.2070& 0.3405\\ 
&CFFT& 0.6518 & 0.4824 & 0.7654 & 0.2034 & 0.4967 & 1934.32 & 0.5324 & 0.6419 \\ \cmidrule{2-10} 
& GRPO & 0.2245& 0.2767& 0.2699 & 0.1297& \underline{0.5222} & \underline{2301.53} & \underline{0.6506}& \underline{0.7410}\\
& DAPO & 0.2431& 0.2798& 0.2687 & 0.1301& 0.5111 & 2330.27 & \textbf{0.6577}& \textbf{0.7423}\\
& GRPON& 0.4437& 0.3656& 0.3189 & 0.1790 & 0.5100 & 2315.97 & 0.6299& 0.7238\\
& SFT $\rightarrow$ GRPON& 0.7501 & 0.5444 & 0.9921 & \textbf{0.2396}& 0.4522 & 901.43 & 0.3012 & 0.4123 \\
& GCPO& 0.7253& 0.5223& 0.9649 & 0.2254 & 0.5011 & 2003.97 & 0.5671& 0.6732\\
& RCPA & 0.7478& 0.5432& 0.9814 & 0.2383& 0.5278 & 2289.18 & 0.6470& 0.7326\\ 
& \cellcolor{gray!30}{S-GRPO(wo CGI)} & \cellcolor{gray!30}{\underline{0.7580}}& \cellcolor{gray!30}{\underline{0.5473}}& \cellcolor{gray!30}{\underline{1.0179}} & \cellcolor{gray!30}{\underline{0.2389}}& \cellcolor{gray!30}{0.5011} & \cellcolor{gray!30}{2209.76} & \cellcolor{gray!30}{0.6067}& \cellcolor{gray!30}{0.7176}\\ 
& \cellcolor{gray!30}{S-GRPO} & \cellcolor{gray!30}{0.7562}& \cellcolor{gray!30}{0.5467}& \cellcolor{gray!30}{1.0106} & \cellcolor{gray!30}{0.2385}& \cellcolor{gray!30}{0.5211}& \cellcolor{gray!30}{2301.18} & \cellcolor{gray!30}{0.6499}& \cellcolor{gray!30}{0.7387}\\ \hline \hline
\multirow{7}{*}{Geo170K}
& BASE & 0.3859& 0.3014& 0.2740 & 0.2901& 0.5122 & 2333.36 & 0.6211& 0.7038\\ \cmidrule{2-10}  
& PEFT & 0.0901& 0.1192& 0.0009 & - &  \textbf{0.6089} &  \textbf{2449.59} & 0.5360& 0.6535\\
& FFT&  \textbf{0.6098}&  \textbf{0.5526}&  \textbf{2.3109} &  \underline{0.5627}& 0.4667 & 2172.37 & 0.5693& 0.6451\\
&CFFT& 0.5719& 0.5091& 1.8827 & 0.5079& 0.5078 & 2199.59 & 0.5888 & 0.6676 \\ \cmidrule{2-10} 
& GRPO & 0.3799& 0.3113& 0.2661 & 0.2878& 0.5022 & \underline{2346.08} & 0.6373& \underline{0.7131}\\
& DAPO & 0.3835& 0.3189& 0.2776 & 0.2989& 0.5111 & 2319.25 & 0.6285& 0.7110\\
& GRPON& 0.4086& 0.3431& 0.3543 & 0.3350 & 0.5122 & 2320.54 & \underline{0.6414}& 0.7062\\
& SFT $\rightarrow$ GRPON& 0.6012 & 0.5504 & 2.2896 & \textbf{0.5629} & 0.4767 & 2199.34 & 0.5801 & 0.6594 \\
& GCPO& 0.5689& 0.5306& 2.0324 & 0.5487 & 0.5089 & 2203.65 & 0.5996& 0.6857\\
& RCPA & 0.5998& 0.5501& 2.2821 & 0.5623& 0.5122 & 2315.37 & 0.6414& 0.7278\\
& \cellcolor{gray!30}{S-GRPO(wo CGI)} & \cellcolor{gray!30}{\underline{0.6086}}& \cellcolor{gray!30}{\underline{0.5525}}& \cellcolor{gray!30}{\underline{2.3087}} & \cellcolor{gray!30}{\underline{0.5627}}& \cellcolor{gray!30}{0.5111} & \cellcolor{gray!30}{2222.08} & \cellcolor{gray!30}{ 0.6231}& \cellcolor{gray!30}{ 0.7098}\\
& \cellcolor{gray!30}{S-GRPO} & \cellcolor{gray!30}{0.6075}& \cellcolor{gray!30}{0.5525}& \cellcolor{gray!30}{2.3008} & \cellcolor{gray!30}{0.5626}& \cellcolor{gray!30}{\underline{0.5211}} & \cellcolor{gray!30}{2322.45} & \cellcolor{gray!30}{ \textbf{0.6430}}& \cellcolor{gray!30}{ \textbf{0.7285}}\\ \hline \hline
\multirow{7}{*}{OpenI}& 
BASE & 0.1155& 0.1299& 0.0002 & 0.0988& 0.5122 & 2333.36 & 0.6211& 0.7038\\ \cmidrule{2-10} 
& PEFT & 0.0786& 0.0977& -& 0.0871& \textbf{0.6111} & \textbf{2433.56} & 0.5508& 0.6631\\
& FFT& \textbf{0.3396}& \textbf{0.2399}& \textbf{0.0903} & \textbf{0.1900}& 0.4111 & 1623.35 & 0.5323& 0.6367\\ 
&CFFT& 0.2698& 0.1889& 0.0813 & 0.1698& 0.4711 & 2100.32& 0.5578& 0.6719\\ \cmidrule{2-10} 
& GRPO & 0.1179& 0.1309& 0.0003 & 0.0994 & 0.5122 & \underline{2356.23} &0.6248& 0.7062\\
& DAPO & 0.1165& 0.1356& 0.0003 & 0.0998 & 0.5111 & 2334.65 & 0.6267& 0.7098\\
& GRPON& 0.1182& 0.1311& 0.0003 & 0.0999& 0.5044 & 2315.37 & 0.6192& \textbf{0.7110}\\
& SFT $\rightarrow$ GRPON& 0.3378& 0.2311& 0.0896 & 0.1839& 0.4311 & 1710.69 & 0.5511& 0.6501\\
& GCPO& 0.2998& 0.2289& 0.0829 & 0.1756& 0.4965 & 2201.43 & 0.5876& 0.6882\\
& RCPA & 0.3342& 0.2325& 0.0886 & 0.1814& 0.5011 & 2321.40 & \underline{0.6285}&0.7062\\
& \cellcolor{gray!30}{S-GRPO(wo CGI)} & \cellcolor{gray!30}{\underline{0.3387}}& \cellcolor{gray!30}{\underline{0.2396}}& \cellcolor{gray!30}{\underline{0.0901}} & \cellcolor{gray!30}{\underline{0.1896}}& \cellcolor{gray!30}{\underline{0.5178}} & \cellcolor{gray!30}{2301.56} & \cellcolor{gray!30}{0.6176}& \cellcolor{gray!30}{0.7001}\\
& \cellcolor{gray!30}{S-GRPO} & \cellcolor{gray!30}{0.3376}& \cellcolor{gray!30}{0.2388}& \cellcolor{gray!30}{0.0899} & \cellcolor{gray!30}{0.1888}& \cellcolor{gray!30}{\underline{0.5211}} & \cellcolor{gray!30}{2345.43} & \cellcolor{gray!30}{\textbf{0.6298}}& \cellcolor{gray!30}{\underline{0.7102}}\\ \bottomrule
\end{tabular}
}}
\caption{Performance comparison of different post-training adaptation methods evaluating both Domain-Specific Ability and General-Purpose Ability. S-GRPO achieves domain-specific performance comparable to, or even exceeding, heavy teacher-forcing methods like Full Fine-Tuning (FFT) and the computationally expensive two-stage pipeline (SFT $\rightarrow$ GRPO), while unlike FFT—which suffers from severe catastrophic forgetting (the "alignment tax")—S-GRPO robustly preserves the pre-trained model's broad multimodal and instruction-following capabilities. Best results are in \textbf{bold}, second-best \underline{underlined}. S-GRPO and its ablation variant are shaded in gray.}
\label{tab: comparison}
\end{table*}

\textbf{Baselines.} We employ Qwen2.5-VL-7B~\citep{Qwen2.5-VL} as our primary base model, chosen for its foundational (not over-optimized) multimodal capabilities and because it has not been pre-aligned using GRPO or similar RL paradigms, ensuring an unbiased testbed. We compare:

(1) \textbf{BASE}: Direct inference with the pre-trained Qwen2.5-VL-7B.

(2) \textbf{SFT-based Methods}: PEFT via LoRA~\citep{Hu2021LoRALA} and Full Fine-tuning (FFT). Drawing on incremental learning~\citep{Kirkpatrick2016OvercomingCF}, we further incorporate a KL-divergence loss into FFT as Continual FFT (CFFT), constraining the fine-tuned output distribution to align with the pre-trained model and mitigating the overwriting of general capabilities.

(3) \textbf{RL-based Methods}: GRPO~\citep{Shao2024DeepSeekMathPT}, DAPO~\citep{Yu2025DAPOAO}, GRPON~\citep{Yan2026ReinforcedCP}, GCPO~\citep{Wu2025GCPOWC}, and RCPA \citep{Yan2026ReinforcedCP}, all evaluated in a strictly RL-driven setting without external "SFT-then-RL" pre-conditioning (the two-stage SFT $\rightarrow$ GRPON pipeline is also reported in Tab. \ref{tab: comparison}). We exclude methods relying on external auxiliary or proprietary LLMs (e.g., Curr-ReFT~\citep{Deng2025BoostingTG}, Scaf-GRPO~\citep{Zhang2025ScafGRPOSG}) to keep comparisons fair.

\begin{table*}[tp]
\centering
{\renewcommand{\arraystretch}{0.85}
\resizebox{0.97\textwidth}{!}{
\begin{tabular}{clcccccccc}
\toprule
\multirow{2}{*}{Datasets} & \multicolumn{1}{c}{\multirow{2}{*}{Methods}} & \multicolumn{4}{c}{Domain-Specific Ability} & \multicolumn{4}{c}{General-Purpose Ability}\\ \cmidrule(r){3-6}\cmidrule(r){7-10} 
& \multicolumn{1}{c}{} & BLEU-1& ROUGE-L & CIDEr & SPICE & MMMU & MME & IFEval-P & IFEval-I \\ \midrule
\multirow{7}{*}{\shortstack{COCO\\Caption}} 
& BASE & 0.4457& 0.3672& 0.2259 & 0.1783& 0.5122 & 2333.36 & 0.6211& 0.7038\\ \cmidrule{2-10} 

& S-GRPO($\delta=0$)& 0.4625& 0.3815& 0.2417 & 0.1908&  \textbf{0.5289} & \textbf{2351.68} & \textbf{0.6587}& \textbf{0.7415}\\

& S-GRPO($\delta=0.3$)& 0.5800& 0.4476& 0.5493 & 0.2099& 0.5222 & 2331.48 & 0.6512& 0.7404\\
& S-GRPO($\delta=0.6$)& 0.6828& 0.5054& 0.8184 & 0.2266& 0.5222 & 2313.80 & 0.6487& 0.7396\\

& \cellcolor{gray!30}{S-GRPO($\delta=0.9$)} & \cellcolor{gray!30}{\underline{0.7562}}& \cellcolor{gray!30}{\underline{0.5467}}& \cellcolor{gray!30}{\underline{1.0106}} & \cellcolor{gray!30}{0.2385}& \cellcolor{gray!30}{0.5211} & \cellcolor{gray!30}{2301.18} & \cellcolor{gray!30}{0.6499}& \cellcolor{gray!30}{0.7387}
\\ 
& S-GRPO($\delta=1$)& \textbf{0.7567}& \textbf{0.5471}& \textbf{1.0170} & \textbf{0.2392} & 0.4889 & 2225.65 & 0.5820& 0.6888\\

\hline 

\end{tabular}
}}
\caption{Ablation study on the control threshold $\delta$ for CGI on the COCO Caption dataset, illustrating the trade-off between domain-specific adaptation and general-purpose capability retention: a higher $\delta$ boosts domain learning but an excessively high value (e.g., $\delta=1$) degrades general capabilities. The highlighted row ($\delta=0.9$) represents the optimal Pareto balance. Best results are in bold, second-best underlined.}
\label{tab: delta}

\end{table*}

\textbf{Evaluation Metrics.} Domain-specific ability is assessed with \textbf{BLEU-1} \citep{papineni-etal-2002-bleu}, \textbf{CIDEr} \citep{Vedantam2014CIDErCI}, \textbf{ROUGE-L} \citep{lin2004rouge}, and \textbf{SPICE} \citep{anderson2016spice}, measuring n-gram overlap, semantic consistency, long-sequence similarity, and structural alignment respectively. General-purpose capability is evaluated using \textbf{MMMU} \citep{Fu2023MMEAC} for cross-domain reasoning, \textbf{MME} \citep{Fu2023MMEAC} for multimodal understanding, and \textbf{IFEval} \citep{Zhou2023InstructionFollowingEF} (IFEval-Prompt/IFEval-Instruct) for instruction-following fidelity.

\textbf{Implementation Details.} We set $\delta=0.9$ in Eq. \ref{eq:reward}, the regularization coefficient $\beta=0.01$ in Eq. \ref{eqn:grpo_obj}, and group size $G=5$. We employ a singular, robust binary verifier (Qwen3-4B-Instruct) that returns $1.0$ for correct direct responses and $0$ otherwise. All experiments are implemented using the EasyR1 RL framework \citep{zheng2025easyr1} and LlamaFactory SFT framework \citep{zheng2024llamafactory}, conducted on a Linux-based server with NVIDIA GPUs.
\begin{figure}[tb]
  \centering
  \includegraphics[width=0.44\textwidth]{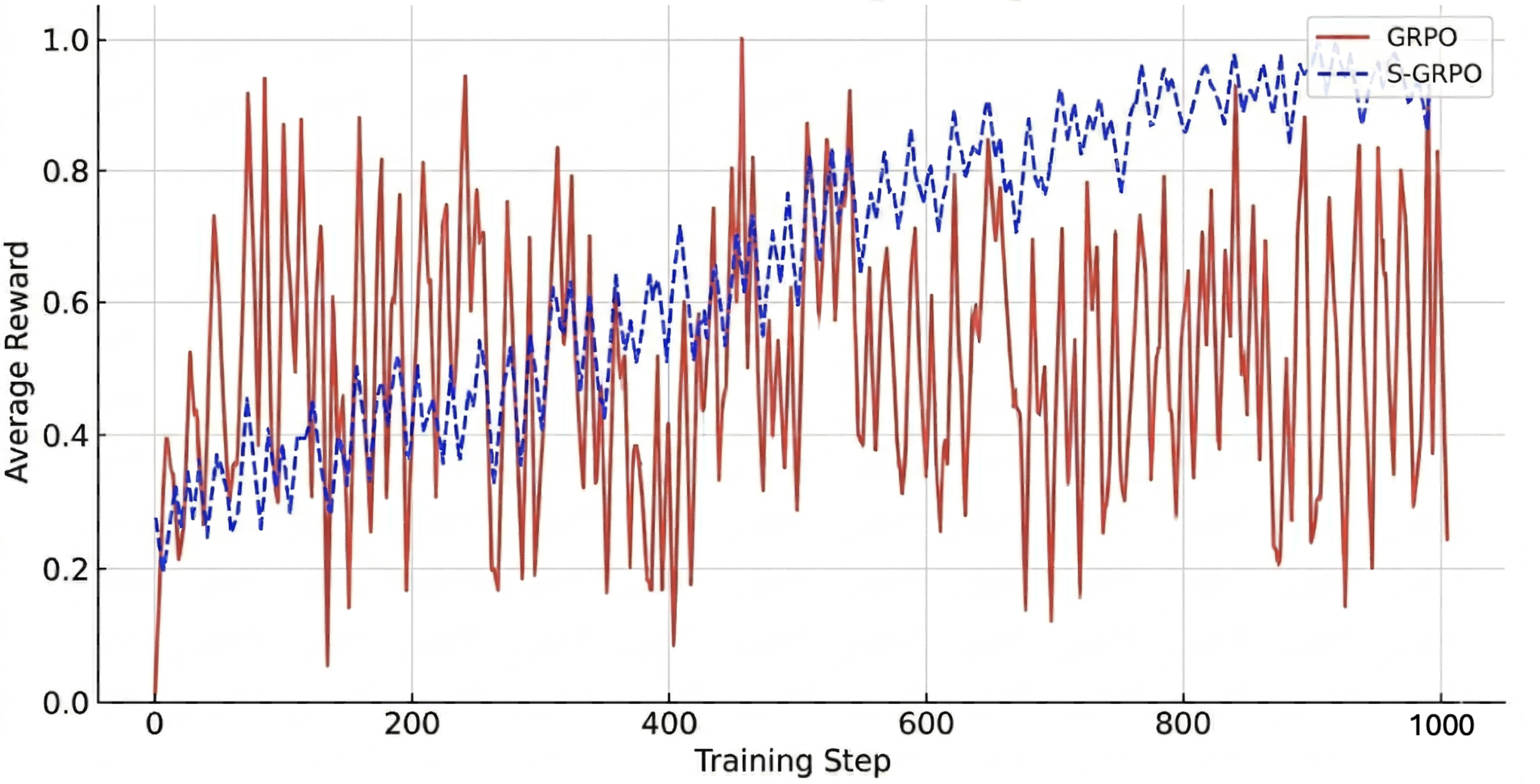}
\caption{Learning dynamics and reward progression of GRPO vs. S-GRPO. Facilitated by a Ground-Truth Injection mechanism, S-GRPO demonstrates a stable, monotonically increasing reward curve leading to robust convergence. In contrast, standard GRPO lacks explicit trajectory guidance and suffers from a severe "flatline" optimization collapse under a strict binary verifier, exhibiting only sporadic, negligible reward spikes lacking meaningful policy gradients.}
  \label{fig:averagereward}
\end{figure}

\subsection{Performance Comparison}
\label{sec: comparison}
Main experimental results are summarized in Tab. \ref{tab: comparison}. S-GRPO delivers superior domain-specific performance while successfully bypassing the alignment tax associated with SFT.

\begin{itemize}[leftmargin=1em]
\item \textbf{FFT Induces Catastrophic Forgetting:} While FFT achieves peak domain scores (e.g., 1.0172 CIDEr on COCO), it severely corrupts general capabilities, with declines of 8.78\% on MMMU, a drop to 735.10 on MME, and over 36\% degradation on IFEval-I.
\item \textbf{Failure of RL-based Methods:} Standard GRPO stalls when applied directly to the base model, yielding a dismal 0.2245 BLEU-1 on COCO under the strict binary verifier, confirming the necessity of our advantage-anchoring mechanism.
\item \textbf{S-GRPO Surpasses Two-Stage Pipelines:} S-GRPO matches or exceeds the two-stage (SFT $\rightarrow$ GRPO) pipeline in domain metrics (1.0106 vs. 0.9921 CIDEr on COCO) without a separate SFT phase, while preserving instruction-following far better (0.7387 IFEval-I).
\end{itemize}
Consistent trends hold on \textbf{Geo170K} (2.3008 CIDEr, comparable to FFT's 2.3109, but better MMMU: 0.5211 vs. 0.4667) and \textbf{OpenI} (0.3376 BLEU-1, 0.2388 ROUGE-L, IFEval-I 0.7102 vs. FFT's 0.6367).

\subsection{Ablation Study}
\label{sec: ablation_study}
\textbf{Effectiveness of Conditional Injection.} We ablate the Ground-Truth Injection mechanism on COCO Caption, as shown in Tab. \ref{tab: comparison}:

\begin{enumerate}
    \item \textbf{No Injection (Standard GRPO)}: Near-zero progress due to severe reward sparsity.
    \item \textbf{Unconditional Injection (S-GRPO wo CGI)}: Injected into every group regardless of success; helps the cold-start problem but introduces a "teacher-forcing" bias that limits autonomous exploration.
    \item \textbf{Conditional Ground-Truth Injection (CGI, Ours)}: Triggers injection only as a fallback, achieving the optimal balance.
\end{enumerate}

\textbf{Reliability of the Verifier.} Evaluated by GPT-5.5, Qwen3-4B-Instruct's semantic judgment accuracy scales robustly with group success rates: \textbf{93.95\%} (0\%), \textbf{94.87\%} (20\%), \textbf{95.76\%} (40\%), \textbf{97.65\%} (60\%), \textbf{99.10\%} (80\%), and \textbf{100\%} (100\% success), confirming its high reliability.

\subsection{Parameter Study}
\label{sec: para_study}
\textbf{Impact of the Control Threshold ($\delta$).} The hyperparameter $\delta$ regulates CGI's strictness. As shown in Tab. \ref{tab: delta}, increasing $\delta$ triggers ground-truth injection more frequently, substantially accelerating domain-specific learning (CIDEr improves from 0.2417 at $\delta=0$ to 1.0170 at $\delta=1$); however, an overly stringent threshold (e.g., $\delta=1$) induces a persistent teacher-forcing effect that degrades general capabilities (MME drops to 2225.65, IFEval-I to 0.6888). Setting $\delta=0.9$ achieves the optimal Pareto balance.
\subsection{Convergence \& Learning Dynamics}
\label{sec: convergence}
Fig. \ref{fig:averagereward} visualizes S-GRPO's learning dynamics: driven by CGI, it exhibits a monotonically increasing reward curve, as high-quality anchors smooth the landscape and guarantee directional gradients during cold-start. In contrast, standard GRPO remains trapped in a zero-gradient plateau, diluted by failed trajectories. As the model's capability improves and CGI intervention naturally diminishes, training transitions smoothly from supervised bootstrapping toward fully autonomous preference optimization.
\section{Conclusion}
We propose S-GRPO, a unified framework harmonizing domain knowledge injection and preference optimization in one on-policy stage. Via CGI, S-GRPO resolves optimization collapse and reward sparsity in cold-start RL, matching FFT while better preserving general LVLM capabilities.

\bibliography{aaai2027}


\end{document}